\documentclass{article}
\usepackage{ijcai22}

\usepackage{times}

\usepackage{soul}
\usepackage{url}
\usepackage[utf8]{inputenc}
\usepackage[small]{caption}
\usepackage{graphicx}
\usepackage{amsmath}
\usepackage{booktabs}
\usepackage{color}

\newcommand\BL[1]{\textcolor{blue}{#1}}

\newcommand\MA[1]{\textcolor{magenta}{#1}}

\usepackage[dvipsnames]{xcolor} 

\usepackage{amsmath}
\usepackage{bm}

\usepackage{dutchcal} 

\usepackage{algorithm}
\usepackage{algpseudocode}

\usepackage{graphicx}
\usepackage{wrapfig} 
\usepackage{amssymb}
\usepackage{tikz} 
\usepackage[T1]{fontenc}

\usepackage{booktabs} 
\usepackage{array}
\usepackage{makecell} 
\usepackage{arydshln} 
\usepackage{multirow}
\usepackage{threeparttable} 
\usepackage{enumitem}

\usepackage[pagebackref=false,breaklinks=true,letterpaper=true,colorlinks,bookmarks=false,citecolor=NavyBlue,linkcolor=red]{hyperref}

\newcommand\bb[1]{\textbf{#1}}
\newcommand\TT[1]{\texttt{#1}}

\usepackage{pifont} 
\newcommand{\cmark}{\ding{51}}%
\newcommand{\xmark}{\ding{55}}%

\newcommand\ie{\textit{i.e.}}
\newcommand\eg{\textit{e.g.}}

\newcommand\vs{\textit{vs.}}
\newcommand\etc{\textit{etc}} 
\usepackage[hang,flushmargin]{footmisc} 

\title{Recent Advances on Neural Network Pruning at Initialization}

\author{
Huan Wang$^{1}$\and
Can Qin$^{1}$\and
Yue Bai$^{1}$\and
Yulun Zhang$^{2}$\And
Yun Fu$^{1}$ \\
\affiliations
\textsuperscript{1}Northeastern University \quad
\textsuperscript{2}ETH Z\"{u}rich
\emails
\{wang.huan, qin.ca, bai.yue\}@northeastern.edu,
yulun100@gmail.com,
yunfu@ece.neu.edu
}

\begin{document}

\maketitle


\begin{abstract}
Neural network pruning typically removes connections or neurons from a \textit{pretrained converged} model; while a new pruning paradigm, \textbf{pruning at initialization} (PaI), attempts to prune a \textit{randomly initialized} network. This paper offers the first survey concentrated on this emerging pruning fashion. 
We first introduce a generic formulation of neural network pruning, followed by the major classic pruning topics. Then, as the main body of this paper, a thorough and structured literature review of PaI methods is presented, consisting of two major tracks (sparse training and sparse selection). Finally, we summarize the surge of PaI compared to PaT and discuss the open problems. Apart from the dedicated literature review, this paper also offers a \href{https://github.com/mingsun-tse/smile-pruning}{code base} for easy sanity-checking and benchmarking of different PaI methods.
\end{abstract}

\section{Introduction}
Network architecture design is a central topic when developing neural networks for various artificial intelligence tasks, especially in deep learning~\cite{LecBenHin15,schmidhuber2015deep}. The wisdom from learning theory suggests that the best generalization comes from a good trade-off between sample size and model complexity~\cite{kearns1994introduction,vapnik2013nature}, which implies to use neural networks of \textit{proper} sizes. However, it is non-trivial to know in practice what size fits properly. \textit{Over-parameterized} networks are thus preferred due to their abundant expressivity when tackling complex real-world problems. In deep learning era, this rule-of-thumb is even more pronounced not only because we are handling more complex problems, but also over-parameterized networks are observed easier to optimize (with proper regularization)~\cite{Simonyan2014Very,resnet} and possibly lead to better generalization~\cite{soltanolkotabi2018theoretical,allen2019learning,zou2020gradient} than their compact counterparts. 

\begin{table}[t]
\centering
\resizebox{\linewidth}{!}{
\setlength{\tabcolsep}{0.6mm}
\begin{tabular}{lcccccccc}
\toprule
\textbf{Pruning paradigm} & \textbf{Weight src} & \textbf{Mask src} & \textbf{Train sparse net?} \\
\midrule
Pruning after training & Convg. net & Convg. net & \cmark~(mostly) \\
\hdashline
\multicolumn{4}{c}{\BL{Pruning at Initialization}} \\
Sparse training (LTH)  & \BL{Init. net}  & Convg. net  & \cmark \\
Sparse training (SNIP)  & \BL{Init. net}  & Init. net  & \cmark \\
Sparse selection (Hidden)    & \BL{Init. net}  & Init. net  & \xmark \\
\bottomrule
\end{tabular}}
\caption{Comparison between~\textbf{pruning at initialization} (PaI) and~\textbf{pruning after training} (PaT). As seen, \textit{weight source} is the axis that differentiates PaI methods from their traditional counterparts: PaI methods inherit weights from a randomly initialized (\textit{Init.}) network instead of a converged (\textit{Convg.}) network. PaI methods can be further classified into two major categories: \textit{sparse training} (picked sparse network will be trained) and \textit{sparse selection} (picked sparse network will \textit{not} be trained). In the parentheses is the representative method of each genre: LTH~\protect\cite{frankle2019lottery}, SNIP~\protect\cite{lee2019snip}, Hidden~\protect\cite{ramanujan2020what}.}
\vspace{-4mm}
\label{tab:differences}
\end{table}

However, over-parameterization brings cost in either testing or training phase, such as excessive model footprint, slow inference speed, extra model transportation and energy consumption. As a remedy, \textbf{neural network pruning} is proposed to remove unnecessary connections or neurons in a neural network without seriously compromising the performance. A typical pruning pipeline comprises 3 steps~\cite{Ree93}: (1) \textbf{pretraining} a (redundant) dense model $\Rightarrow$ (2) \textbf{pruning} the dense model to a sparse one  $\Rightarrow$ (3) \textbf{finetuning} the sparse model to regain performance. Namely, pruning is considered as a \textit{post-processing} solution to fix the side-effect brought by pretraining the dense model. 
This post-processing paradigm of pruning has been practiced for more than 30 years and well covered by many surveys, \eg, a relatively outdated survey~\cite{Ree93}, recent surveys of pruning alone~\cite{gale2019state,blalock2020state,hoefler2021sparsity} or as a sub-topic under the umbrella of model compression and acceleration~\cite{sze2017efficient,cheng2018recent,cheng2018model,deng2020model}.

However, a newly surging pruning paradigm, \textbf{pruning at initialization} (PaI), is absent from these surveys (we are aware that PaI is discussed in one very recent survey~\cite{hoefler2021sparsity}, yet merely with one subsection; our paper aims to offer a \textit{more thorough and concentrated} coverage). Unlike \textit{pruning after training} (PaT) (see Tab.~\ref{tab:differences}), PaI prunes a \textit{randomly initialized} dense network instead of a pretrained one. PaI aims to train (or purely select) a sparse network out of a randomly initialized dense network to achieve (close to) \textit{full accuracy} (the accuracy reached by the dense network). This had been believed unpromising as plenty of prior works have observed training the sparse network from scratch underperforms PaT, until recently, lottery ticket hypothesis (LTH)~\cite{frankle2019lottery} and SNIP~\cite{lee2019snip} successfully find sparse networks which can be trained from scratch to full accuracy. They open new doors to efficient sparse network training with possibly less cost. 

This paper aims to present a comprehensive coverage of this emerging pruning paradigm, discussing its historical origin, the status quo, and possible future directions. The rest of this paper is organized as follows. First, Sec.~\ref{sec:background} introduces the background of network pruning, consisting of a generic formulation of pruning and the classic topics in PaT. Next, Sec.~\ref{sec:PaI} presents the thorough coverage of PaI methods. Then, Sec.~\ref{sec:open} summarizes the PaI fashion and discusses the open problems. Finally, Sec.~\ref{sec:conclusion} concludes this paper.

\section{Background of Neural Network Pruning} \label{sec:background}
\subsection{A Generic Formulation of Pruning} \label{subsec:pruning_formulation}
The stochastic gradient descent (SGD) learner of a neural network parameterized by $\mathbf{w}$ produces a model \textit{sequence} which finally converges to a model with desired performance:
\begin{equation}
    \big\{\mathbf{w}^{(0)}, \mathbf{w}^{(1)},~\cdots, \mathbf{w}^{(k)},~\cdots,  \mathbf{w}^{(K)} \big\},
\end{equation}
where $K$ denotes the total number of training iterations. In PaT, we only need the \textit{last} model checkpoint ($k=K$) as the base model for pruning. However, the new pruning paradigm asks for models at different iterations, \eg, LTH needs the model at iteration $0$~\cite{frankle2019lottery} (or an early iteration~\cite{frankle2020linear}, we consider this also as $0$ for simplicity). To accommodate these new cases, we allow a pruning algorithm to have access to the \emph{whole model sequence} (which we can easily cache during the model training). Then, pruning can be defined as a \textit{function}, which takes the \textit{model sequence} as input (along with the training dataset $\mathcal{D}$), and outputs a pruned model $\mathbf{w}^{\ast}$.

\begin{figure*}
    \centering
    \includegraphics[width=\linewidth]{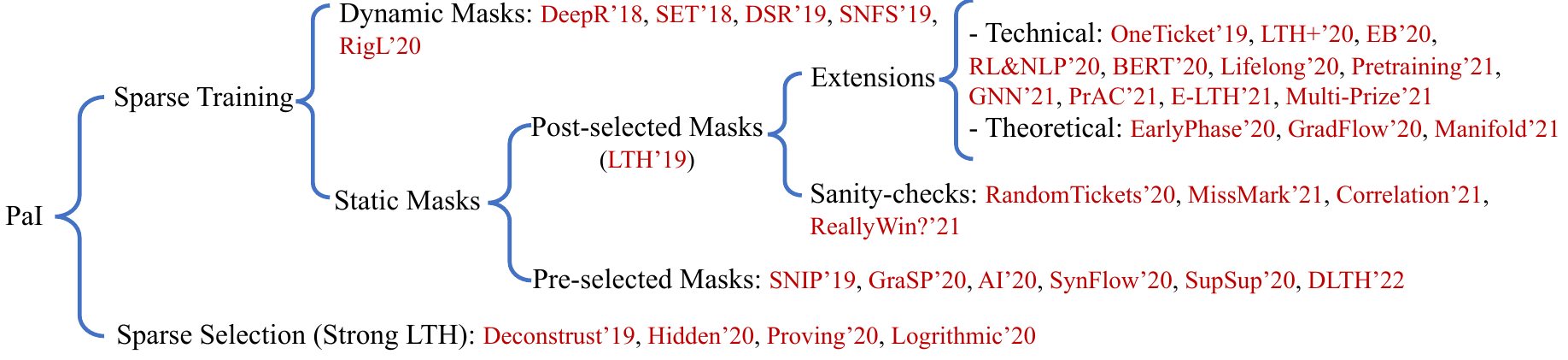}
    \vspace{-5mm}
    \caption{Overview of pruning at initialization (PaI) approaches, classified into two general groups: \textit{sparse training} and \textit{sparse selection}. For readability, references are omitted in this figure but the paper abbreviations (right beside the abbreviation is the year when the paper appeared). Please see Sec.~\ref{sec:PaI} for detailed introductions of them. Due to limited length, this paper only outlines the primary methods, see \textit{full} collection at \MA{https://github.com/mingsun-tse/awesome-pruning-at-initialization}.}
    \label{fig:overview}
    \vspace{-3mm}
\end{figure*}

We need two pieces of information to make a neural network exact, its topology and the associated parameter values. \textbf{(1)} In the case of a sparse network, the sparse topology is typically defined by a mask tensor (denoted by \bb{m}, with the same shape as $\mathbf{w}$), which can be obtained via a function, $\bb{m} = \mathbf{f_1}(\mathbf{w}^{(k_1)}; \mathcal{D})$, that is, $\mathbf{f_1}$ can utilize the data available to help decide the masks (\eg, in regularization-based pruning methods, such as~\cite{wang2021neural}). \textbf{(2)} Meanwhile, the weights in the pruned model can be adjusted from their original values to mitigate the incurred damage~\cite{OBS,wang2019eigendamage,wang2018structured,wang2021neural}. This step can be modeled as another function $\mathbf{f}_2(\mathbf{w}^{(k_2)}; \mathcal{D})$. Together, pruning can be formulated as 
\begin{equation}
\begin{split}
    \mathbf{w'} &= \mathbf{f_1}(\mathbf{w}^{(k_1)}; \mathcal{D}) \odot \mathbf{f_2}(\mathbf{w}^{(k_2)}; \mathcal{D}), \\
    \mathbf{w^\ast} &= \mathbf{f_3}(\mathbf{w'}; \mathcal{D}),
\end{split}
\label{eq:pruning_formulation}
\end{equation}
where $\odot$ represents the Hadamard (element-wise) product; $\mathbf{f_3}$ models the finetuning process (which is omitted in the sparse selection PaI methods). Note that, the input models for $\mathbf{f_1}(*)$ and $\mathbf{f_2}(*)$ can be from \emph{different} iterations. By this definition, the pruning paradigms in Tab.~\ref{tab:differences} can be specified as follows,
\begin{itemize}
    \item Pruning after training (PaT): $k_1=K, k_2=K$;
    \item Sparse training: $k_1=K$ (LTH) or $0$ (SNIP), $k_2=0$, and $\mathbf{f_2} = \mathbf{I}$ (identity function), $\mathbf{f_3} \neq \mathbf{I}$;
    \item Sparse selection: $k_1=k_2=0$, and $\mathbf{f_2}=\mathbf{f_3}=\mathbf{I}$.  
\end{itemize}

\subsection{Classic Topics in Pruning} \label{subsec:traditional_pruning}


There are mainly four critical questions we need to ask when pruning a specific model: \textit{what} to prune, \textit{how many} (connections or neurons) to prune, \textit{which} to prune exactly, and \textit{how} to schedule the pruning process, corresponding to four classic topics in pruning as follows. We only briefly explain them here as they are \textit{not} our main focus. For a more comprehensive coverage, we refer the readers to~\cite{hoefler2021sparsity}.

\vspace{0.2em}
\noindent \bb{(1)~Sparsity structure}. Weights can be pruned in patterns. The shape of the pattern, named \textit{sparsity structure}, decides the basic pruning element of a pruning algorithm. The smallest structure, of course, is a single weight element, \ie, no structure at all. This kind of pruning is called \emph{unstructured pruning}. The pruning pattern larger than a single weight element can be called \emph{structured pruning} in general. Although there are different levels of granularity when defining structured pruning (see~\cite{mao2017exploring}), structured pruning typically means filter or channel pruning in the literature.

Practically, unstructured pruning is mainly favored for model size \textit{compression}, structured pruning more favored for \textit{acceleration} due to the hardware-friendly sparsity structure. Acceleration is more imperative than compression for modern deep networks, so most PaT works focus on structured pruning currently. We will see that, differently, most PaI methods focus on unstructured pruning instead.

\vspace{0.2em}
\noindent \bb{(2)~Pruning ratio}. Pruning ratios indicate how many weights to remove. In general, there are two ways to determine pruning ratios. \textbf{(i)} The first is to \emph{pre-define} them. Namely, we know exactly how many parameters will be pruned before the algorithm actually runs. This scheme can be further specified into two sub-schemes. One is to set a \emph{global} pruning ratio (\ie, how many weights will be pruned for the whole network); the other is to set \emph{layer-wise} pruning ratios. \textbf{(ii)} The second is to decide the pruning ratio by other means. This way mostly appears in the regularization-based pruning methods, which remove weights by driving them towards zero via penalty terms. A larger regularization factor typically leads to more sparsity, \ie, a larger pruning ratio. However, how to set a proper factor to achieve the desired sparsity usually demands heavy tuning. Several methods have been proposed to improve this~\cite{wang2018structured,wang2021neural} -- the pruning ratios are usually pre-specified. Recent years also have seen some works that automatically search the optimal layer-wise pruning ratio~\cite{he2018amc}. No consensus has been reached on which is better. 


\vspace{0.2em}
\noindent \bb{(3)~Pruning criterion}. Pruning criterion decides which weights to remove given a sparsity budget. It is considered one of the most critical problems in network pruning, thus has received the most research attention so far. The most simple criterion is weight magnitude (or equivalently, $L_1$-norm for a tensor)~\cite{han2015learning,li2017pruning}. Because of its simplicity, it is the most prevailing criterion in PaT now. We will see it also being widely used in PaI.

Albeit much exploration in this topic, there are no criteria that prove to be significantly better than the others (actually, the simple magnitude pruning has been argued as (one of) the SOTA~\cite{gale2019state}). Since this topic is already well discussed in prior surveys in PaT, we will not cover it in depth here. One point worth mention is that, the core idea underpinning most  pruning criteria now is to \emph{select the weights whose absence induces the least loss change}. This idea and its variant have been followed for a long time in PaT~\cite{OBD,OBS,wang2019eigendamage,MolTyrKar17,molchanov2019importance}, which continues to PaI, as we will see.

\vspace{0.2em}
\noindent \bb{(4)~Pruning schedule}. After the three aspects determined above, we finally need to specify the schedule of pruning. There are three typical choices~\cite{wang2019structured}. \textbf{(i)} \textit{One-shot}: network sparsity (defined by the ratio of zeroed weights in a network) goes from $0$ to a target number \emph{in a single step}, then finetune. \textbf{(ii)} \textit{Progressive}: network sparsity goes from $0$ to a target number \emph{gradually}, typically along with network training; then finetune. \textbf{(iii)} \textit{Iterative}: network sparsity goes from $0$ to an intermediate target number, then finetune; then repeat the process until the target sparsity is achieved. Note, both (ii) and (iii) are characterized by \emph{pruning interleaved with network training}. Therefore, there is \emph{no} fundamental boundary between the two. Some works thus use the two terms interchangeably. One consensus is that, progressive and iterative pruning outperform the one-shot counterpart when pruning the same number of weights because they allow more time for the network to adapt. In PaI, LTH employs \textit{iterative} pruning, which increases the training cost significantly. PaI works featured by pre-selected masks (\eg,~\cite{wang2020picking}) are thereby 
motivated to resolve this problem.


\section{Pruning at Initialization (PaI)} \label{sec:PaI}
\subsection{Overview: History Sketch}
\vspace{0.2em}
\noindent \bb{Debut: LTH and SNIP}. The major motivation of PaI \vs~PaT is to achieve \textit{cheaper and simpler} pruning -- pruning a network at initialization prior to training ``\textit{eliminates the need for both pretraining and the complex pruning schedule}'' (as quoted from SNIP~\cite{lee2019snip}). Specially, two works lead this surge, LTH~\cite{frankle2019lottery} and SNIP. Importantly, both make a similar claim: Non-trivially sparse networks can be trained to full accuracy in isolation, which was believed barely feasible before. Differently, LTH selects masks from a \textit{pretrained} model while SNIP selects masks from the \textit{initialized} model, \ie, post-selected \vs~pre-selected masks, as shown in Fig.~\ref{fig:overview}. 

\vspace{0.2em}
\noindent \bb{Follow-ups of LTH and SNIP}. \textbf{(1)} There are two major lines of LTH follow-ups. Acknowledging the efficacy of LTH, one line is to expand its universe, \eg, scaling LTH to larger datasets and models (\eg, ResNet50 on ImageNet), validating it on non-vision domains (\eg, natural language processing), and proposing theoretical foundations. The other line takes a decent grain of salt about the efficacy of LTH, keeps sanity-checking. \textbf{(2)} At the same time, for the direction led by SNIP, more methods come out focusing on better pruning criteria.

\vspace{0.2em}
\noindent \bb{Dynamic masks}. The masks in both LTH and SNIP are \textit{static}, namely, they are \textit{fixed} during training. Some researchers (such as \cite{evci2020rigging}) conjecture that dynamic and adaptive masks during training may be better, thus introduce another group of methods featured by \textit{dynamic masks}. Together, the static-and dynamic-mask methods complete the realm of \textit{sparse training}.

\vspace{0.2em}
\noindent \bb{Sparse selection}. When some researchers attempt to understand LTH through empirical studies, they discover an interesting phenomenon. \cite{zhou2019deconstructing} surprisingly find that the random network selected by winning ticket in LTH actually has non-trivial accuracy \textit{without any training}. This discovery brings us strong LTH: Sparse subnet picked from a dense network can achieve full accuracy \textit{even without further training}, and opens a new direction named \textit{sparse selection}.

The current PaI universe is primarily made up of the above two categories, sparse training and sparse selection. A method tree overview of PaI is presented in Fig.~\ref{fig:overview}. Next, we elaborate the major approaches in each genre in length.

\subsection{Sparse Training} \label{subsec:sparse_training}
\noindent \bb{Static masks: post-selected}. Sparse training methods of this group are pioneered by \TT{LTH}~\cite{frankle2019lottery}. The pruning pipeline in LTH has three steps: First, a randomly initialized network is trained to convergence; Second, employ magnitude pruning to obtain the masks, which defines the topology of the subnet; Third, apply the masks to the \textit{initial} network to obtain a subnet and train the subnet to convergence. This process can be repeated iteratively (\ie, iterative magnitude pruning, or IMP). The authors surprisingly find the subnet can achieve comparable (or even better occasionally) accuracy to the dense network. The \textit{lottery ticket hypothesis} is thus proposed: ``\textit{dense, randomly-initialized, feed-forward networks contain subnetworks (winning tickets) that—when trained in isolation— reach test accuracy comparable to the original network in a similar number of iterations}.''

Follow-up works of LTH mainly fall into two groups. One group identifies the legitimacy of LTH and attempts to seek a broader application or understanding of it. The other group proposes several sanity-check ablations of LTH, taking a decent grain of salt with its validity. These two groups are thus summarized as \textit{extensions} and \textit{sanity-checks} in Fig.~\ref{fig:overview}.

\vspace{0.2em}
\noindent \textit{(1) Extensions}. The original LTH is only validated on small datasets (MNIST and CIFAR-10). \cite{frankle2020linear} (\TT{LTH+}) later generalize LTH to ResNet50 on ImageNet by applying the masks not to the initial weights but to the weights after a few epochs of training.   
\cite{morcos2019one} (\TT{OneTicket}) discover that winning tickets can generalize across a variety of datasets and optimizers within the natural images domain. Besides, winning tickets generated with larger datasets consistently transfer better than those generated with smaller datasets, suggesting that winning tickets contain inductive biases generic to neural networks.
\cite{yu2020playing} (\TT{RL\&NLP}) find the lottery ticket phenomenon also presents on RL and NLP tasks and use it to train compressed Transformers to high performance. \cite{chen2020lottery} validates lottery ticket phenomenon on \TT{BERT} subnetworks at 40\% to 90\% sparsity. 
\cite{chen2020long} (\TT{Lifelong}) introduce two pruning methods, bottom-up and top-down, to find winning tickets in the lifelong learning scenario.
\cite{chen2021lottery} (\TT{Pretraining}) locate matching subnetworks at 59.04\% to 96.48\% sparsity that transfer universally to multiple downstream tasks with no performance degradation for supervised and self-supervised pre-training.
\cite{chen2021unified} (\TT{GNN}) present a unified graph neural network (GNN) sparsification framework that simultaneously prunes the graph adjacency matrix and the model weights and generalizes LTH to GNN for the first time.
The iterative train-prune-retrain cycles on full training set in LTH can be very expensive. To resolve this, \cite{zhang2021efficient} introduce Pruning Aware Critical (\TT{PrAC}) set, which is a \textit{subset} of training data. PrAC set takes only 35.32\% to 78.19\% of the full set (CIFAR10, CIFAR100, Tiny ImageNet) with similar effect, saving up to 60\% to 90\% training iterations. Moreover, they find the PrAC set can generalize \textit{across different network architectures}. 
Similarly, \TT{E-LTH}~\cite{chen2021elastic} also attempt to find winning tickets generic to different networks without the expensive IMP process. They achieve so by transforming winning tickets found in one network to another deeper or shallower one \textit{from the same family}.
\cite{you2020drawing} find that winning tickets can be identified at the very early training stage (thus they term the tickets \textit{early-bird}, or \TT{EB}, tickets) via low-cost training schemes (\eg, early stopping and low-precision training) at large learning rates, achieving up to 4.7$\times$ energy saving while maintaining the performance.
%
%
\cite{diffenderfer2021multi} (\TT{Multi-Prize}) find winning tickets on \textit{binary} neural networks for the first time.

Beside the technical extensions of LTH above, researchers also try to understand LTH more theoretically. \cite{frankle2020early} (\TT{EarlyPhase}) analyze the early phase of deep network training. \cite{evci2020gradient} (\TT{GradFlow}) present a gradient flow perspective to explain why LTH happens. \cite{zhang2021validating} (\TT{Manifold}) verify the validity of LTH by leveraging dynamical systems theory and inertial manifold theory. More works on the LTH theories study a \textit{stronger} version of LTH, \ie, sparse selection. We thus defer them to Sec.~\ref{subsec:sparse_selection}.

In short, extensions of LTH mainly focus on LTH\textit{+X} (\ie, generalizing LTH to other tasks or learning settings, \etc.), identifying cheaper tickets, understanding LTH better.

\vspace{0.2em}
\noindent \textit{(2) Sanity-checks}. The validation of LTH seriously hinges on experimental setting, which actually has been controversial since the very beginning of LTH. In the same venue where LTH is published, another pruning paper~\cite{liu2019rethinking} actually draws a pretty different conclusion from LTH. They argue that a subnetwork with \textit{random initialization} (\vs~the winning tickets picked by IMP in LTH) can be trained from scratch to match pruning a pretrained model. Besides,~\cite{gale2019state} also report that they cannot reproduce LTH. The reasons, noted by~\cite{liu2019rethinking}, may reside in the experimental settings -- \cite{liu2019rethinking} focus on filter pruning and use momentum SGD with a large initial learning rate (LR) (0.1), while~\cite{frankle2019lottery} tackle unstructured pruning and ``mostly uses Adam~\cite{kingma2014adam} with much lower learning rates''. This debate continues on. \cite{su2020sanity} (\TT{RandomTickets}) notice that \textit{randomly} changing the preserved weights in each layer (the resulted initializations are thus named \textit{random tickets}), while keeping the layer-wise pruning ratio, does not affect the final performance, thus introduce the \textit{random tickets}. \cite{frankle2021pruning} (\TT{MissMark}) later report similar observation. This property ``suggests broader challenges with the underlying pruning heuristics, the desire to prune at initialization, or both''~\cite{frankle2021pruning}. Meanwhile, \cite{liu2021lottery} (\TT{Correlation}) find that there is a strong correlation between initialized weights and the final weights in LTH, when the LR is not sufficiently large, thereby they argue that ``the existence of winning property is correlated with an insufficient DNN pretraining, and is unlikely to occur for a well-trained DNN''. This resonates with the conjecture by \cite{liu2019rethinking} above that learning rate is a critical factor making their results seemingly contradicted with LTH. \cite{ma2021sanity} (\TT{ReallyWin?}) follow up this direction and present more concrete evidence to clarify whether the winning ticket exists across the major DNN architectures and/or applications.

\vspace{0.2em}
\noindent \bb{Static masks: Pre-selected}. SNIP~\cite{lee2019snip} pioneers the direction featured by pre-selected masks. \TT{SNIP} proposes a pruning criterion named \emph{connectivity sensitivity} to select weights based on a straightforward idea of loss preservation, \ie, removing the weights whose absence leads to the least loss change. After initialization, each weight can be assigned a score with the above pruning criterion, then remove the last-$p$ fraction ($p$ is the desired pruning ratio) parameters for subsequent training. Later, \cite{wang2020picking} argue that it is the training dynamics rather than the loss value itself that matters more at the beginning of training. Therefore they propose gradient signal preservation (\TT{GraSP}) in contrast to the previous loss preservation, based on Hessian. 
Concurrently to GraSP~\cite{wang2020picking}, \cite{lee2020signal} seek to explain the feasibility of SNIP through the lens of signal propagation. They empirically find pruning damages the dynamical isometry~\cite{saxe2014exact} of neural networks and propose an data-independent initialization, \textit{approximated isometry} (\TT{AI}), which is an extension of exact isometry to sparse networks. 
Later, \TT{SynFlow}~\cite{tanaka2020pruning} proposes a new data-independent criterion, which a variant of magnitude pruning, yet taking into account the interaction of different layers.
Meanwhile, \TT{SupSup}~\cite{wortsman2020supermasks} proposes a training method to select different masks (supermasks) for thousands of tasks in continual learning, from a \textit{single} fixed random base network.
Another very recent static-mask work is \TT{DLTH} (dual LTH)~\cite{bai2022dual}. In LTH, initial weights are given, the problem is to find the matching masks, while in DLTH, the masks are given, the problem is to find the matching initial weights. Specifically, they employ a growing $L_2$ regularization~\cite{wang2021neural} technique to transform the original random weights to the desired condition.

To summarize, similar to PaT, the primary research attention in this line of PaI works also lies in \textit{pruning criterion}. Actually, many PaI works propose \textit{pretty similar} criterion formulas to those in PaT (see Tab.~\ref{tab:pruning_criterion}).

\vspace{0.2em}
\noindent \bb{Dynamic masks}. Another sparse training track allows the masks to be changed during training. 
\TT{DeepR}~\cite{bellec2018deep} employs dynamic sparse parameterization with stochastic parameter updates for training.
\TT{SET}~\cite{mocanu2018scalable} adopts magnitude pruning and random growth for dynamic network topology optimization.
\TT{DSR}~\cite{mostafa2019parameter} adaptively allocates layer-wise sparsity ratio with no need of manual setting and shows training-time structural exploration is necessary for best generalization.
\TT{SNFS}~\cite{dettmers2019sparse} proposes to employ the momentum of gradients to bring back pruned weights.
\TT{RigL}~\cite{evci2020rigging} also uses magnitudes for pruning, yet they employ the absolute gradients for weight growing.

\begin{table}[t]
    \centering
    \resizebox{0.9\linewidth}{!}{
    \setlength{\tabcolsep}{4mm}
    \renewcommand{\arraystretch}{1.1}
    \begin{tabular}{lcc}
    \toprule
    Method      & Pruning criterion \\
    \midrule 
    Skeletonization (1989) & $-\nabla_{\mathbf{w}}\mathcal{L} \odot \mathbf{w}$\\
    OBD (1990)             & $\text{diag}(H)\mathbf{w} \odot \mathbf{w}$ \\
    Taylor-FO (2019)  & $(\nabla_{\mathbf{w}}\mathcal{L} \odot \mathbf{w})^2$\\
    \hdashline
    SNIP (2019)    & $|\nabla_{\mathbf{w}}\mathcal{L} \odot \mathbf{w}|$ \\
    GraSP (2020)   & $-H\nabla_{\mathbf{w}}\mathcal{L} \odot \mathbf{w}$ \\
    SynFlow (2020) & $\frac{\partial \mathcal{R}}{\partial \mathbf{w}} \odot \mathbf{w}, \mathcal{R} = \mathbf{1}^\top (\Pi_{l=1}^L |\mathbf{w}^{[l]}|) \mathbf{1}$ \\
    \bottomrule
    \end{tabular}}
    \vspace{-1mm}
    \caption{Summary of pruning criteria in \textit{static-mask} sparse training methods. Above the dash line are PaT methods. $\mathcal{L}$ denotes the loss function; $H$ represents Hessian; $\mathbf{1}$ is the all ones vector; $l$ denotes the $l$-th layer of all $L$ layers. Skeletonization~\protect\cite{mozer1989skeletonization}. Taylor-FO~\protect\cite{molchanov2019importance}.}
    \label{tab:pruning_criterion}
    \vspace{-3mm}
\end{table}

The essential idea of dynamic masks is to enable zeroed weights to rejoin the training again. Of note, this idea is not newly brought up in PaI. It has been explored too in pruning after training. In terms of the criteria, notably, because the masks need to re-evaluated \textit{frequently} during training in dynamic-mask methods, the pruning and growing criteria cannot be costly. Therefore, \textit{all} these criteria are based on \textit{magnitudes or gradients} (see Tab.~\ref{tab:criterion_dynamic_masks}), which are readily available during SGD training.

\vspace{0.2em}
\subsection{Sparse Selection} \label{subsec:sparse_selection}
Sparse training methods still need to optimize the values of the subnet after picking it out of the dense model. Some other works have found another optimization scheme: instead of optimizing the weight values, \emph{optimize the network topology}. Namely, when a network is randomly initialized, all the values of each connection are fixed. The goal is to find a subnet from the dense network \emph{without further training the subnet}. This line of works 
is firstly pioneered by~\cite{zhou2019deconstructing} (\TT{Deconstruct})  where they try to understand the mysteries of LTH. They find the subnet picked by LTH can achieve non-trivial accuracies already (without training). This implies that, although the original full network is randomly initialized, the chosen subnet is not really random. The subnet picking process itself serves as a kind of training. Therefore, \cite{zhou2019deconstructing} propose the notion of \emph{supermasks} (or masking as training) along with a proposed algorithm to optimize the masks in order to find better supermasks.

\begin{table}[t]
    \centering
    \resizebox{0.9\linewidth}{!}{
    \renewcommand{\arraystretch}{1.1}
    \setlength{\tabcolsep}{1mm}
    \begin{tabular}{lcccc}
    \toprule
    Method      & Pruning criterion & Growing criterion \\
    \midrule
    DNS (2016)             & $|\mathbf{w}|$ & $|\mathbf{w}|$ (pruned weights also updated)\\
    \hdashline
    DeepR (2018)           & stochastic & random \\
    SET (2018)             & $|\mathbf{w}|$ & random\\
    DSR (2019)             & $|\mathbf{w}|$ & random \\
    SNFS (2020)            & $|\mathbf{w}|$ & momentum of $\nabla_{\mathbf{w}}\mathcal{L}$ \\
    RigL (2020)            & $|\mathbf{w}|$ & $|\nabla_{\mathbf{w}}\mathcal{L}|$ \\ 
    \bottomrule
    \end{tabular}}
    \vspace{-1mm}
    \caption{Summary of pruning and growing criteria in \textit{dynamic-mask} sparse training methods. Above the dash line are the PaT methods. DNS~\protect\cite{guo2016dynamic}; see other references in Sec.~\ref{subsec:sparse_selection}.}
    \label{tab:criterion_dynamic_masks}
    \vspace{-3mm}
\end{table}

The method in \cite{zhou2019deconstructing} is only evaluated on small-scale datasets (MNIST and CIFAR). Later, \cite{ramanujan2020what} (\TT{Hidden}) introduce a trainable score for each weight and update the score to minimize the loss function. The trainable scores are eventually used to decide the selected subnetwork topology. The method achieves strong performance for the first time. \textit{E.g.}, they manage to pick a subnet out of a \emph{random} Wide ResNet50. The subnet is smaller than ResNet34 while delivers better top-1 accuracy than the trained ResNet34 on ImageNet. This discovery is summarized as a stronger version of LTH, \textit{strong LTH}: ``\textit{within a sufficiently over-parameterized neural network with random weights (e.g. at initialization), there exists a subnet- work that achieves competitive accuracy}''~\cite{ramanujan2020what}.

The above are the breakthroughs in terms of technical algorithms. In terms of theoretical progress, \cite{malach2020proving} (\TT{Proving}) propose the theoretical basis of the strong LTH, suggesting that ``pruning a randomly initialized network is as strong as optimizing the value of the weights''. Less favorably, they make assumptions about the norms of the inputs and of the weights. Later, \cite{orseau2020logarithmic,pensia2020optimal} (\TT{Logarithmic}) remove the aforementioned limiting assumptions while providing significantly \textit{tighter} bounds: ``the over-parameterized network only needs a \textit{logarithmic} factor (in all variables but depth) number of neurons per weight of the target subnetwork''. \cite{diffenderfer2021multi} have shown that scaled \textit{binary} networks can be pruned to approximate any target function, where the optimization is \textit{polynomial} in the width, depth, and approximation error, akin to~\cite{malach2020proving}. \cite{sreenivasan2021finding} further demonstrate that a \textit{logarithmically} over-parameterized binary network is enough to approximate any target function, similar to the advance of~\cite{orseau2020logarithmic,pensia2020optimal} over~\cite{malach2020proving}.

\section{Summary and Open Problems} \label{sec:open}
\subsection{Summary of Pruning at Initialization}
Easily seen, the biggest paradigm shift from PaT to PaI is to use a randomly initialized model to inherit weights for the sparse network, just as the PaI term (pruning at \textit{initialization}) suggested. The primary goal is to achieve efficient \textit{training}, in contrast to efficient \textit{inference} aimed by PaT. This section answers what are the similarities and major differences between PaI~\vs~the traditional fashion in a big-picture level.

\vspace{0.2em}
\noindent \bb{Sparse training: Much overlap between PaI and PaT}. Not surprisingly, there is considerable overlap between PaI and PaT. \textbf{(1)} For LTH-like works, the adopted pruning scheme is mainly (iterative) magnitude pruning, which has been developed for more than thirty years and widely considered as the most simple pruning method. \textbf{(2)} For pre-selected-mask methods, the proposed pruning criteria for PaI are reminiscent of those proposed in PaT (Tab.~\ref{tab:pruning_criterion}). Therefore, we may continue seeing the wisdom in PaT being transferred to PaI.

In terms of the four primary aspects in pruning (Sec.~\ref{subsec:traditional_pruning}), PaI does not bring much change to the pruning ratio, criterion, and schedule. However, for \textit{sparsity structure}, it does have a significant impact. First, structured pruning (or filter pruning) itself is not much interesting in the context of PaI because given the layer-wise sparsity ratios filter pruning at initialization reduces to \textit{training from scratch} a dense network (with a smaller width). Then the major technical problem is how to initialize a slimmer network properly, which has been well-explored. This is probably why most PaI works focus on unstructured pruning rather than structured pruning. Second, the well-known LTH is proposed for unstructured pruning. As far as we know, to date, \textit{no} work has validated the hypothesis with filter pruning. The reason behind is still elusive.

\vspace{0.2em}
\noindent \bb{Sparse selection: Significant advance of PaI~\vs~PaT}.  
The common belief behind PaT is that the remaining parameters possess knowledge learned by the original redundant model. Inheriting the knowledge is better than starting over, which has been also empirically justified by many works and thus become a common practice for more than thirty years. In stark contrast, PaI inherits the parameter values right from the beginning. It is tempting to ask: \textit{At such an early phase of neural network training, do the weights really possess enough knowledge (to provide base models for subsequent pruning)}? Or more fundamentally, is neural network training essentially to learn the knowledge \emph{from nothing} or to reveal the knowledge \emph{the model already has}? PaT implies the former, while PaI (especially sparse selection~\cite{ramanujan2020what,malach2020proving}) suggests the latter. This, we believe, is (arguably) the most profound advance PaI has brought about. The understanding of this question will not only bestow us practical benefits, but also, more importantly, a deeper theoretical understanding of deep neural networks.

\subsection{Open Problems}
\noindent \bb{Same problems as pruning after training}. Given the overlap between PaI and PaT, nearly all the open problems in PaT will also apply to PaI, too, centering on the four classic topics in pruning: how to find better sparsity structure being hardware friendly and meanwhile optimization friendly, how to find better pruning criteria, how to allocate the sparsity more properly cross different layers, and how to schedule the sparsification process more wisely, in the hopes of higher performance given the same sparsity budget. We do not reiterate these topics here considering they have been well treated in existing pruning surveys (\eg,~\cite{hoefler2021sparsity}).

\vspace{0.2em}
\noindent \bb{Under-performance of PaI}. The idea of PaI is intriguing, however, in terms of practical performance, PaI methods still \emph{underperform} PaT methods by an obvious margin. \textit{E.g.}, according to the experiments in~\cite{wang2020picking}, with VGG19 and ResNet32 networks on CIFAR-10/100, both SNIP and GraSP are \emph{consistently} outperformed across different sparsity levels by two traditional pruning methods (OBD~\cite{OBD} and MLPrune~\cite{zeng2019mlprune}), which are not even close to the state-of-the-art. \cite{frankle2021pruning} also report similar observation. In this sense, there is still a long road ahead before we can really ``save resources at training time''~\cite{wang2020picking}.

\vspace{0.2em}
\noindent \bb{Under-development of sparse libraries}. Despite the promising potential of sparse training, the practical acceleration benefit to date has not been satisfactory. \textit{E.g.}, SNFS~\cite{dettmers2019sparse} claims ``up to 5.61x faster training'', yet in practice, due to under-development of sparse matrix multiplication libraries, this benefit cannot be materialized at present. To our best knowledge, \textit{few} works have reported \textit{wall-time speedup} in sparse training. Development of sparse training libraries thus can be a worthy future cause. In this regard, notably, dynamic-mask methods pose even more severe issues than static-mask methods as the constantly changing masks will make it harder for the hardware to gain acceleration. If the training is not really getting faster, this may fundamentally undermine the motivation of sparse training, which actually is the main force of PaI now.

\textit{To sum, the major challenges facing PaI is to deliver the \textbf{practical training speedup} with no (serious) performance compromised, as it promises.} This hinges on a more profound comprehension of the performance gap between sparse training and PaT, as well as the advance of sparse matrix libraries.

\vspace{0.2em}
\noindent \bb{Code base}. As discussed, PaI works (especially LTH-like) severely hinge on \textit{empirical} studies. However, it is non-trivial in deep learning to tune hyper-parameters. In this paper, we thereby also offer a code base\footnote{\MA{https://github.com/mingsun-tse/smile-pruning}} with systematic support of popular networks, datasets, and logging tools, in the hopes of aiding researchers and practitioner focus more on the insights and core methods instead of the tedious engineering details.

\section{Conclusion} \label{sec:conclusion}
This paper presents the first survey concentrated on neural network pruning at initialization (PaI). The road map of PaI~\vs~its pruning-after-training counterpart is studied with a thorough and structured literature review. We close by outlining the open problems and offer a code base towards easy sanity-checking and benchmarking of different PaI methods.

\vspace{0.3em}
\noindent \bb{Acknowledgments}. We thank Jonathan Frankle, Alex Renda, and Michael Carbin from MIT for their very helpful suggestions to our work.

\newpage
\small
\bibliographystyle{named}
\bibliography{ijcai22}
\end{document}